\pdfoutput=1

\documentclass[11pt]{article}

\usepackage[]{acl}

\usepackage{times}
\usepackage{latexsym}
\usepackage{amsfonts} 
\usepackage{booktabs}
\usepackage{mathtools}
\usepackage{float}
\usepackage{adjustbox}
\usepackage{xcolor}
\usepackage{multirow}

\usepackage[T1]{fontenc}

\usepackage[utf8]{inputenc}

\usepackage{microtype}
\usepackage{graphicx}

\title{Multiformer: A Head-Configurable Transformer-Based Model \\
for Direct Speech Translation}

\author{Gerard Sant, 
  Gerard I. Gállego,
  Belen Alastruey \and
  Marta R. Costa-Jussà\\
  TALP Research Center, Universitat Politècnica de Catalunya, Barcelona \\
  \texttt{gerard.muniesa@estudiantat.upc.edu} \\ \texttt{\{gerard.ion.gallego, belen.alastruey, marta.ruiz\}@upc.edu}}

\begin{document}
\maketitle
\begin{abstract}

Transformer-based models have been achieving state-of-the-art results in several fields of Natural Language Processing. However, its direct application to speech tasks is not trivial. The nature of this sequences carries problems such as long sequence lengths and redundancy between adjacent tokens. Therefore, we believe that regular self-attention mechanism might not be well suited for it. 

Different approaches have been proposed to overcome these problems, such as the use of efficient attention mechanisms. However, the use of these methods usually comes with a cost, which is a performance reduction caused by information loss. In this study, we present the Multiformer, a Transformer-based model which allows the use of different attention mechanisms on each head. By doing this, the model is able to bias the self-attention towards the extraction of more diverse token interactions, and the information loss is reduced. Finally, we perform an analysis of the head contributions, and we observe that those architectures where all heads relevance is uniformly distributed obtain better results. Our results show that mixing attention patterns along the different heads and layers outperforms our baseline by up to $0.7$ BLEU.

\end{abstract}

\section{Introduction}

Conventionally, Speech-to-text Translation (ST) task has been addressed through cascade approaches \citep{ney1999speech}, which consists of the concatenation of an Automatic Speech Recognition block (ASR), for the audio transcription, with another Machine Translation block (MT), for the translation of such transcription into the desired language. However, this approach ignores some information present in the audio, since it translates from the audio transcript, and is also vulnerable to error propagation, since an error in the ASR block automatically causes a mistranslation \citep{sperber-paulik-2020-speech,bentivogli-etal-2021-cascade}. Consequently, end-to-end alternatives based on an encoder-decoder structure and attention mechanisms have become increasingly popular in recent years \citep{anastasopoulos-etal-2016-unsupervised, duong-etal-2016-attentional, weiss2017sequence}. These are capable of translating the audio without the explicit need for transcription, thus avoiding the problems of the cascade approach and allowing unified optimization of the training parameters.

The advent of the Transformer \citep{vaswani2017attention} revolutionized the MT field, enabling models based on this architecture to achieve the state-of-the-art results. Nowadays, Transformer-based models are used to process all types of data, such as images \citep{parmar2018image} or speech \citep{dong_speech_transformer, di2019adapting}. However, due to its self-attention mechanism, the vanilla Transformer scales quadratically with the input sequence length, which makes it extremely inefficient when processing long sequences.

In speech tasks, it is common to extract audio features every 10 ms to build the input sequences, which causes them to be considerably longer than text sequences. Moreover, since the representation of a single phoneme requires several tokens \citep{igras2013length, inproceedingslength}, the presence of redundancy among the audio tokens is inferred. Therefore, state-of-the-art architectures propose the implementation of down sampling strategies prior to the model collapsing adjacent vectors in a fixed way \citep{berard2018end, di-gangi-etal-2019-enhancing, wang-etal-2020-fairseq}. Similarly, some studies propose to extract more informative sequences using pretrained compression modules \citep{salesky-etal-2019-exploring, zhang-etal-2020-adaptive, gaido-etal-2021-ctc}, obtaining considerable translation quality gains. While these achieve good results, we propose another approach, questioning the use of multi-head self-attention (MHSA), originally proposed for text, for the information extraction from speech sequences.

The closest work to ours was done by \citet{alastruey2021efficient}, who used Longformer's \citep{beltagy2020longformer} local attention pattern as an efficient alternative to self-attention for speech processing. However, they observed that, due to the scarcity of global context in the encoder output, the quality of the translations was slightly hindered. Recently, inspired by Linformer's \citep{wang2020linformer} attention, \citet{papi-etal-2021-speechformer} proposed ConvAttention as an attention mechanism that, by compressing keys and values, is more efficient and therefore able to directly process long sequences. However, this mechanism is not used as a replacement of the encoder self-attention, but as an extra input processing before a CTC-based compression module \citep{gaido-etal-2021-ctc}.

Our contribution to ST field is a new Transformer variant, the Multiformer, an architecture based on the S2T Transformer by \citet{wang-etal-2020-fairseq}. Our architecture enables the use of different attention mechanisms in the same encoder layer, by configuring individually the pattern of each head. With this approach, the Multiformer is able to apply efficient attention mechanisms, while maintaining the ability to learn both local and global content from speech sequences. This diversity among heads in a layer is also meant to stimulate a more varied information extraction and, therefore, reduce the presence of low-relevant heads \citep{voita-etal-2019-analyzing, michel2019sixteen, bian-etal-2021-attention, zhang-etal-2021-enlivening}. Furthermore, we explore the use of different head configurations for each encoder layer. This could help to adapt the attention mechanisms to the needs of each layer. To the best of our knowledge, this is the first study that allows this kind of head-wise configuration.




\section{Model}

In this section, we first introduce a new self-attention module that allows the use of multiple attention mechanisms in parallel (\S \ref{cap:MHMA}). Next, we explain the Multiformer (\S \ref{cap:Multiformer}), which replaces the Transformer encoder MHSA by the new proposed module. 

\subsection{Multi-head Multi-attention}\label{cap:MHMA}

An increasing number of studies have observed the presence of redundant heads in multi-head self-attention \citep{michel2019sixteen, bian-etal-2021-attention, zhang-etal-2021-enlivening}. Moreover, \citet{voita-etal-2019-analyzing} even tried to prune them, and observed that the quality of the translations (in MT) was almost not affected. This suggests that the model does not exploit the full potential present in the use of attention heads. In addition, the quadratic time and memory complexity of Self-Attention with respect to the input sequence length makes it impossible to use it directly in Speech tasks. To address this challenge, end-to-end ST models are based on reducing the length of speech sequences, usually by a factor of 4, through compression techniques, so that they can be processed by the Transformer \citep{di-gangi-etal-2019-enhancing, wang-etal-2020-fairseq}. However, after this compression, the resulting sequences are still considerably longer and more redundant than their text counterparts. \citet{alastruey2021efficient} proposed the use of efficient Transformers for ST, but, as observed in different tasks by \citet{tay2020long}, they suffer from a drop in performance quality. The main reason for this deterioration is that most efficient Transformers propose strategies that deprive the model of the ability to learn all types of content from the input stream.\footnote{In efficient Transformers that approximate the $softmax()$ function \citep{choromanski2020rethinking} or the attention matrix \citep{wang2020linformer}, the quality drop can be attributed to an imperfection in these approximations.} 


\begin{figure}[h]
    \centering
\includegraphics[scale=0.13]{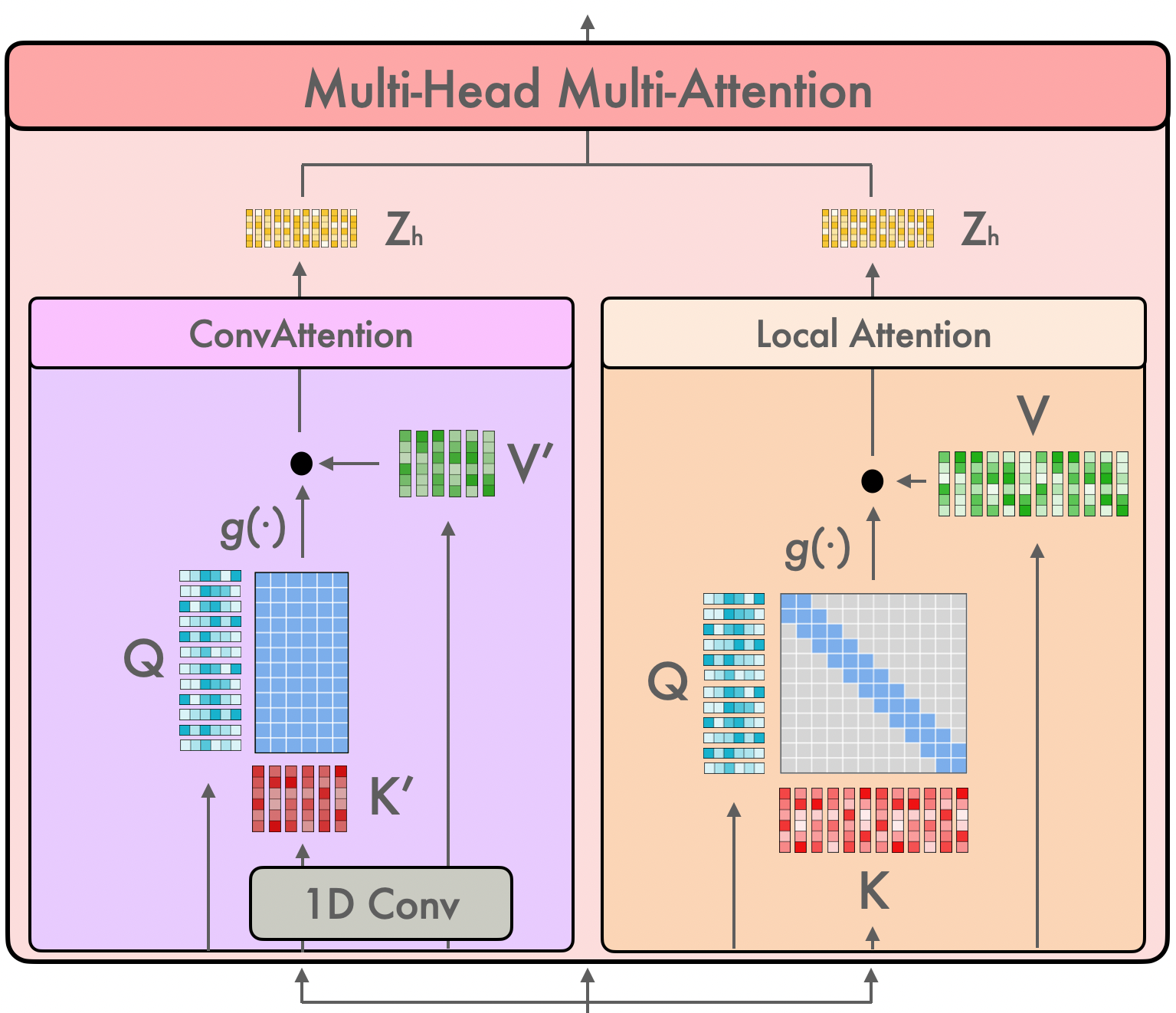}
    \caption{Scheme of the MHMA with the representation of each of the attention mechanisms it incorporates. The function $g()$ denotes the downscaling and computation of the $softmax()$ function.}
    \label{fig:MHMA}
\end{figure}

To solve the aforementioned problems, we propose multi-head multi-attention (MHMA) (Figure \ref{fig:MHMA}) which, using heads with different attention mechanisms, is meant to force a more diversified learning, thus (i) hindering the presence of irrelevant heads and (ii) allowing the model to learn both local and global content from the input sequence, while applying efficient attention mechanisms. The MHMA module is manually set by selecting the type of attention mechanism for each head in each layer within the following ones:

\paragraph{ConvAttention.} Efficient attention mechanism proposed by \citet{papi-etal-2021-speechformer}. The ConvAttention compresses the keys and values by means of a convolutional layer, decreasing the size of the attention matrix by a factor of $\chi$, to reduce the original complexity to O($(\frac{n}{\chi})^2$). By not compressing the queries, they manage to maintain the dimensions of the input sequence at the output.

\paragraph{Local Attention.} Attention mechanism with a sliding window pattern \citep{beltagy2020longformer}. It only computes the product between queries and keys of nearby tokens within the same input sequence, so it is more efficient than the regular Self-Attention. In particular, given a fixed window size $w$, each token attends to $\frac{w}{2}$ tokens on each side, achieving a linear scaling (O($n\times w$)) of the module complexity. As in \citet{alastruey2021efficient}, this attention pattern is intended to force the learning of local relations, while being more efficient.




\subsection{Multiformer}\label{cap:Multiformer}

The Multiformer is a Transformer-based architecture inspired by \citet{wang-etal-2020-fairseq}. The original model consists on a regular Transformer, preceded by two 1D convolutional layers, that help to tackle speech-specific problems such as a longer sequence length or information redundancy in adjacent tokens. The Multiformer proposes to modify the self-attention module on each encoder layer by a MHMA, since we believe that this module could be helpful to deal with speech.

The introduction of MHMA allows the model to learn from different representational and contextual levels. This enables the construction of architectures capable of extracting different kinds of information from the input sequence, while performing more efficient attention mechanisms. In addition, the model is biased towards learning different types of token interactions, hindering the presence of irrelevant heads. 

\begin{figure}[h]
    \centering
    \includegraphics[scale=0.27]{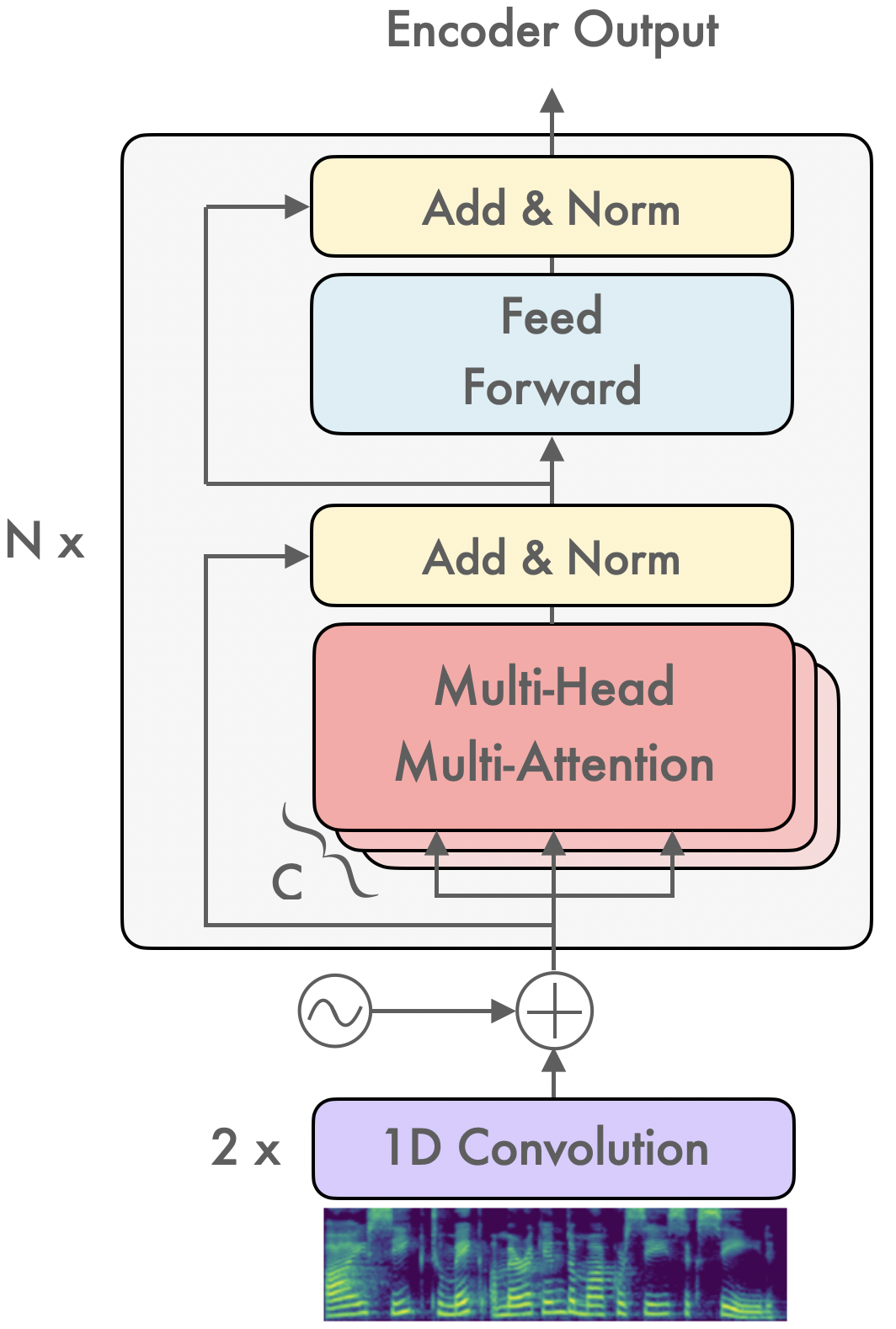}
    \caption{Diagram of the Multiformer encoder. It comprises N layers, each one with C heads that can use different attention mechanisms.}
    \label{fig:Multiformer}
\end{figure}



However, the generation of attention diversity at the head level does not address the presence of redundancy between layers noted by \citet{dalvi-etal-2020-analyzing}, who, using linear Center Kernel Alignment \citep{kornblith2019similarity}, observed that, except for the last two layers, layer redundancy increases throughout the encoder. Moreover, the information processed by each layer differs, hence using the same MHMA configuration in all encoder layers may not be the optimal.

Therefore, the Multiformer (Figure \ref{fig:Multiformer}) allows the use of different MHMA configurations, which is meant to create architectures that process the speech sequence in a more progressive manner. This approach emphasizes the learning of different content along the encoder layers, while hampering information redundancy.

\begin{figure*}[h]
    \centering
\includegraphics[scale=0.185]{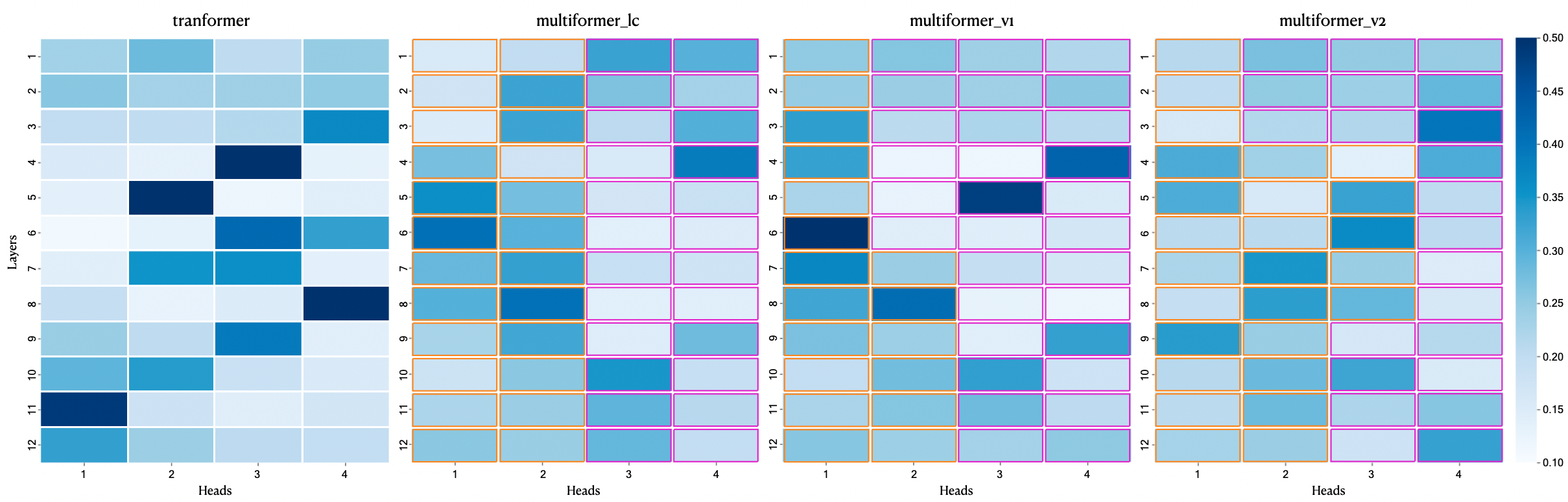}
    \caption{Head relevance for each layer of the proposed models. They have been computed using the median of $n$ contributions (equation \ref{eqn:head_contribution}) from 500 random samples of the en-de training partition. Heads marked in \textcolor{orange}{orange} use \textcolor{orange}{Local Attention} while those in \textcolor{blue!20!magenta}{purple} are using \textcolor{blue!20!magenta}{ConvAttention}.}
    \label{fig:Analisis_heads}
\end{figure*}

\section{Heads Contribution Analysis}\label{AnalsisTeorico}
MHMA allows the use of different attention mechanisms in parallel, therefore we wanted to evaluate the head contribution in each of the encoder layers. 


In general, given an input sequence of $n$ tokens $\{\mathbf{x_{1}},...,\mathbf{x_{n}}\}\in \mathbb{R}^{n\times d}$ and a model with an embeddings dimension $d$ (head dimension $d_h$), the output of each attention-head $\mathbf{z^{h}_{i}}\in \mathbb{R}^{d_h}$  is:

\begin{equation}{
    \mathbf{z}_{i}^{h} = \sum_{j}^{N}\mathbf{A}_{i,j}^h \mathbf{Wv}^{h} \; \mathbf{x}_j 
    }
    \label{eqn:heads}
\end{equation}



where $\mathbf{A}_{i,j}^{h}$ is the attention weight of token $j$ on token $i$ and $\mathbf{Wv}^{h}\in \mathbb{R}^{d_h\times d}$ is the learned projection matrix of the values. The final output representation of the attention module $\mathbf{y_{i}} \in \mathbb{R}^{d}$ is:

\begin{equation}{
    \mathbf{y}_{i}= \mathbf{Wo} \; \text{Concat}\{ \mathbf{z}_{i}^{1}, ..., \mathbf{z}_{i}^{H} \} + \mathbf{b}_o
    }\label{eqn:attention_out}
\end{equation}



with $\mathbf{Wo} \in \mathbb{R}^{d \times H\cdot d_h}$ as the output projection matrix trained jointly with the model, and $\mathbf{b}_o \in \mathbb{R}^{d}$ referring to the bias. As previously done for interpretability research \citep{kobayashi-etal-2020-attention}, the above expression (equation \ref{eqn:attention_out}) can be rewritten as follows:


\begin{equation}{
    \mathbf{y}_{i} = \sum_{h}^{H}\mathbf{Wo}^{h} \; \mathbf{z}_{i}^{h} + \mathbf{b}_o
    }
    \label{eqn:output_layer_descomposed}
\end{equation}

where $\mathbf{Wo}^{h} \in \mathbb{R}^{d \times d_h}$ is the part of the output projection matrix corresponding to each head.
Note that from this last expression, it can be defined $\mathbf{\xi}^{h}_i = \mathbf{Wo}^{h}\mathbf{z}_{i}^{h}\in \mathbb{R}^{d}$ as the projected output vector of a head. 

Inspired by \citet{kobayashi-etal-2020-attention}, for each layer, we define the contribution of each head to the attention output $\mathbf{y}_{i}$ as the Euclidean norm of the projected output vector of heads: 

\begin{equation}{
    {c}_{i,h} = {\rvert \rvert \mathbf{\xi}^{h}_i \lvert \lvert}_2
    }
    \label{eqn:head_contribution}
\end{equation}







\section{Experiments}
In this section we first explain the training details (\S \ref{Settings}) in order to ensure reproducibility of experiments.\footnote{Code available: \url{https://github.com/mt-upc/fairseq/tree/multiformer}} Then we briefly describe Multiformer architectures and the procedure we followed (\S \ref{descripcion}). 

\subsection{Experimental Settings}\label{Settings}

The Multiformer architectures have been trained on 3 different language directions of the MuST-C dataset \citep{cattoni2021must}. This corpus consists of audio, transcriptions, and translations of TED talks in English. MuST-C provides 8 language directions ranging in size from 385 (Portuguese) to 504 (Spanish) hours of transcribed and translated speech. 

To ensure a faithful comparison with the baseline model, the small architecture of the S2T Transformer in Fairseq \citep{wang-etal-2020-fairseq}, all our models consist of 12 encoder layers, 6 decoder layers and 4 heads in each attention layer. The embedding dimension of the model is 256. Moreover, following the baseline architecture, we have kept the convolutional layers with downsampling prior to the model.


For the ConvAttention, we use a kernel size of $5$ and a stride of $2$, reducing the length of keys and values to the half. Regarding the Local Attention, as in \citet{alastruey2021efficient} we have chosen a window size of $64$ tokens. These hyperparameters have been employed in all Multiformer architectures. For a detailed description of training parameters, see appendix \ref{sec:DetailExp}.



\begin{table*}[t]
\centering
\small \begin{tabular}{l@{\hskip 4pt}| @{\hskip 7pt} c @{\hskip 5pt} c @{\hskip 5pt} c @{\hskip 4pt}| @{\hskip 7pt} l @{\hskip 5pt} c @{\hskip 5pt} c@{\hskip 4pt}| @{\hskip 7pt} l @{\hskip 5pt} c @{\hskip 5pt} c @{\hskip 4pt}| @{\hskip 4pt} c}
\hline
\multirow{2}{*}{\textbf{Model}} & \multicolumn{3}{c}{\textbf{en-de \space \space \space \space \space \space \space \space}} & \multicolumn{3}{c}{\textbf{en-fr \space \space \space \space \space \space \space \space}} & \multicolumn{3}{c}{\textbf{en-es}} & \multirow{2}{*}{\textbf{Avg}$(\Delta\%)$}\\
  & BLEU & $\Delta$BLEU & $\Delta\%$& BLEU & $\Delta$BLEU & $\Delta\%$ & BLEU  & $\Delta$BLEU & $\Delta\%$ \\
\hline
baseline & $22.65$ & - & - & $32.97$ & - & - & $26.99$ & - & - & - \\
\hline
local\_attention & $22.69$ & $+0.04$ & $+0.17$ & $33.00$ & $+0.03$ & $+0.09$ & $27.10$ & $+0.11$ & $+0.41$ & $+0.22$ \\
conv\_attention & $22.45$ & $-0.20$ & $-0.88$ & $33.07$ & $+0.10$ & $+0.30$ & $26.96$ & $-0.04$ & $-0.15$ & $-0.73$  \\
\hline
multiformer\_lc & $22.80$ &$+0.15$& $+0.66$ & $33.25$ &$+0.28$& $+0.85$ & $27.56$ & $+0.57$ & $+2.11$ & $+1.21$ \\
multiformer\_v1 & $\bf{23.16}$ &$\bf{+0.51}$& $\bf{+2.25}$& $33.10$& $+0.13$& $+0.39$& $\bf{27.68}$&$\bf{+0.69}$& $\bf{+2.56}$ & $\bf{+1.73}$\\
multiformer\_v2 & $22.98$ &$+0.33$& $+1.46$& $\bf{33.26}$ &$\bf{+0.29}$& $\bf{+0.88}$& $27.44$& $+0.45$& $+1.67$& $+1.34$ \\
\hline
\end{tabular}
\caption{\label{results}
BLEU results in 3 different language directions of the MuST-C dataset, English$\rightarrow$German (en-de), English$\rightarrow$French (en-fr) and English$\rightarrow$Spanish (en-es). Relative improvements are calculated with respect to the baseline \citep{wang-etal-2020-fairseq}.}
\end{table*}

\subsection{Experiments Description}\label{descripcion}

First, we trained two architectures based on a single attention mechanism (Local or ConvAttention), in order to obtain a comparison between models with and without diversity.

After this, we trained the first Multiformer architecture, the $multiformer\_lc$. It has a configuration of the MHMA with 2 heads of ConvAttention and 2 heads of Local Attention for all encoder layers. Then, we analyzed the contribution of each head following the methodology described in \S\ref{AnalsisTeorico}. This allowed us to better understand the needs of each layer, and to propose architectures based on this. From Figure \ref{fig:Analisis_heads}, it can be seen that in the first 3 layers, the $multiformer\_lc$ assigns low relevance to the representations extracted by one of the Local Attention heads, which could indicate the prioritization of the global context in the first layers. In the middle layers, a change in this trend is observed, with Local Attention heads acquiring more importance. As for the last layers, we see an equal relevance distribution between heads of both mechanisms.

These observations have motivated the training of the $multiformer\_v1$, which tries to correct the abandonment of Local Attention heads observed in the initial layers. It consists of substituting a Local Attention head for a ConvAttention head in the first six layers of the encoder.

Finally, the $multiformer\_v2$ is built more strictly from the analysis. It incorporates 3 different MHMA configurations. In the first 3 layers, it uses 1 head of Local Attention and 3 heads of ConvAttention. The next 5 layers (from the 4th to the 8th) use 3 Local Attention heads and 1 ConvAttention head, to finish the remaining 4 layers with 2 heads of each type.

In general, it is clear that, whereas the baseline uses few heads in most layers, Multiformer architectures\footnote{More details in Table \ref{tab:Arquitecturas} in the Appendices.} force the model to have a more uniformly distributed contribution between heads.



\section{Results}

First, it can be observed from Table \ref{results}, that the efficient architecture based only on Local Attention ($local\_attention$) already obtains the same results as the baseline, suggesting the presence of unnecessary computations in self-attention. Unlike previous works \citep{alastruey2021efficient}, this architecture maintains the convolutional layers, so the amount of global content within the attention mechanism is higher using the same window size. On the other hand, while the architecture based exclusively on ConvAttention ($conv\_attention$), manages to achieve baseline results in English$\rightarrow$French (en-fr) and English$\rightarrow$Spanish (en-es), its score in English$\rightarrow$German (en-de) drops $0.2$ BLEU, suggesting the need for a higher resolution extraction of representations for that language pair. 


Secondly, analyzing the heads contribution of the baseline architecture, we can observe that the heads contribution tends to accumulate in few heads. This means we obtain similar conclusions than \citet{voita-etal-2019-analyzing}, but for the ST setting. 
Furthermore, our work goes one step further, showing that those architectures where the heads contribution is uniformly distributed, obtain a higher performance. This finding confirms that, in ST, some heads on a regular Transformer tend to learn irrelevant information. This shows that MHSA might not be as capable as expected of extracting different kinds of patterns, unless it is biased on purpose towards doing so.

In particular, all Multiformer variants improve the results obtained by the baseline and the $local\_attention$ and $conv\_attention$ architectures. 
However, these improvements are not equal in all languages pairs, and go from $0.15$ to $0.57$ BLEU for $multiformer\_lc$, from $0.13$ to $0.69$ BLEU for $multiformer\_v1$ and from $0.29$ to $0.45$ BLEU for $multiformer\_v2$, becoming the latter the architecture with the most robust gains. 

\section{Conclusions}

In this paper, we present the Multiformer, the first Transformer-based model that allows to combine different attention mechanisms in the MHSA module. This helps the model extracting different types of token interactions from each head, hence preventing the appearance of irrelevant heads. By applying this diversity of attention patterns with efficient mechanisms, the model is able to maintain both local and global context across encoder layers while being more efficient. Experiments on 3 language pairs show that all Multiformer architectures outperform the results achieved by the S2T Transformer in the ST task, with an improvement up to $0.69$ BLEU for the English-Spanish direction in the $multiformer\_v1$.

\section{Acknowledgements}

This work was partially funded by the ADAVOICE project, PID2019-107579RB-I00 / AEI / 10.13039/501100011033, and the UPC INIREC scholarship nº3522. We would like to thank Javier Ferrando for his advice on the heads contribution analysis.




\bibliographystyle{acl_natbib}

\appendix
\renewcommand{\arraystretch}{1.7}
\begin{table*}[t]
    \centering
    \begin{tabular}{l | l | l | l}
    \toprule
    \multicolumn{1}{c}{\small \textbf{Name}} &
    \multicolumn{1}{c}{\small \textbf{}} &
    \multicolumn{1}{c}{\small \textbf{MHMA Configurations}} &
    \multicolumn{1}{c}{\small \textbf{}} \\
    \midrule
    \hline

    \small conv\_attention &\small  $ 12 \times 
        \begin{psmallmatrix}
            4 \times & Conv\, (5,\, 2)
        \end{psmallmatrix}$  &\small  \space \space \space \space \space \space \space \space \space \space \space \space \space \space \space \space \space \space \space - &\small  \space \space \space \space \space \space \space \space \space \space \space \space \space \space \space \space \space -  \\
        \hline
    \small local\_attention &\small  $ 12 \times 
        \begin{psmallmatrix}
            4 \times & Local\, (64)
        \end{psmallmatrix}$  &\small  \space \space \space \space \space \space \space \space \space \space \space \space \space \space \space \space \space \space \space -  &\small \space \space \space \space \space \space \space \space \space \space \space \space \space \space \space \space \space -  \\
        \hline
    \small multiformer\_lc &\small  $ 12 \times 
        \begin{psmallmatrix}
            2 \times & Local\, (64)\\
            2 \times & Conv\, (5,\, 2)
        \end{psmallmatrix}$  &\small  \space \space \space \space \space \space \space \space \space \space \space \space \space \space \space \space \space \space \space -  &\small \space \space \space \space \space \space \space \space \space \space \space \space \space \space \space \space \space -  \\
        \hline
    \small multiformer\_v1 &\small $ 6 \times 
        \begin{psmallmatrix}
            1 \times & Local\, (64)\\
            3 \times & Conv\, (5,\, 2)
        \end{psmallmatrix} $ &\small  $ 6 \times 
        \begin{psmallmatrix}
            2 \times & Local\, (64)\\
            2 \times & Conv\, (5,\, 2)
        \end{psmallmatrix}$ &\small \space \space \space \space \space \space \space \space \space \space \space \space \space \space \space \space \space  - \\
        \hline
    \small multiformer\_v2 &\small $ 3 \times 
        \begin{psmallmatrix}
            1 \times & Local\, (64)\\
            3 \times & Conv\, (5,\, 2)
        \end{psmallmatrix} $ &\small  $ 5 \times 
        \begin{psmallmatrix}
            3 \times & Local\, (64)\\
            1 \times & Conv\, (5,\, 2)
        \end{psmallmatrix} $ &\small  $ 4 \times 
        \begin{psmallmatrix}
            2 \times & Local\, (64)\\
            2 \times & Conv\, (5,\, 2)
        \end{psmallmatrix}$ \\
    \bottomrule
    \end{tabular}
    \centering
    \captionsetup{justification=centering}
    \caption[Trained Multiformer architectures]{Multiformer architectures. The notation for each configuration is as follows: $ N_{layers} \times 
        \begin{psmallmatrix}
            N_{heads} & \times & Attention\, (hyperparameters)\\
        \end{psmallmatrix} $.}
    \label{tab:Arquitecturas}
\end{table*}
\newpage
\section{Detailed Experimental Settings}
\label{sec:DetailExp}

The training has been performed using the label smoothed cross entropy loss \citep{Szegedy_2016_CVPR_smoth} and the Adam optimizer \citep{kingma2015adam}. The learning rate has been set to $2\cdot 10^{-3}$ with an inverse square-root scheduler and 10,000 warm-up updates. We have set a maximum number of 32,000 tokens for the construction of the mini-batches and an update frequency of 5. The training has been hosted on 2 NVIDIA GeForce RTX 2080 Ti GPUs until the completion of 50,000 updates. For a better performance in ST, models have been pretrained in ASR \citep{berard2018end}.\footnote{For the use of the ASR pretrained encoder in ST training, the best checkpoint has been used, being this the one that obtains the lowest loss.} For this pretraining, all the parameters have been set as in ST, with the exception of the learning rate, which has been set to $1\cdot 10^{-3}$.

For the S2T evaluation of the architectures, we averaged the 7 checkpoints around the best one and then computed the BLEU score \citep{papineni-etal-2002-bleu}.

To visualize the layer-level relevance of each head (Figure \ref{fig:Analisis_heads}), we computed the median of the contributions (\S \ref{AnalsisTeorico}) of each head for all the tokens in 500 random samples. Since we want to observe which head is the most relevant during training, the en-de training partition was used.





\end{document}